\def\year{2019}\relax
\documentclass[letterpaper]{article} %DO NOT CHANGE THIS
\usepackage{aaai19}  %Required
\usepackage{times}  %Required
\usepackage{helvet}  %Required
\usepackage{courier}  %Required
\usepackage{url}  %Required
\usepackage{graphicx}  %Required
\usepackage{stfloats}
\usepackage{subfigure}
\usepackage{algorithm}
\usepackage{algorithmic}
\usepackage{amsfonts}
\usepackage{amsmath}
\usepackage{multirow}
\begin{document}
\title{Visualized Insights into the Optimization Landscape of \\
Fully Convolutional Networks}
\author{Jianjie Lu,$^{1}$ Kai-yu Tong$^{2}$\\
The Chinese University of Hong Kong\\
1 Department of Electronic Engineering, jacklu@link.cuhk.edu.hk\\
2 Department of Biomedical Engineering, kytong@cuhk.edu.hk\\
Shatin, NT, Hong Kong SAR, China, 999077\\
}
\maketitle
\begin{abstract}
Many image processing tasks involve image-to-image mapping, which can be addressed well by fully convolutional networks (FCN) without any heavy preprocessing. Although empirically designing and training FCNs can achieve satisfactory results, reasons for the improvement in performance are slightly ambiguous. Our study is to make progress in understanding their generalization abilities through visualizing the optimization landscapes. The visualization of objective functions is obtained by choosing a solution and projecting its vicinity onto a 3D space. We compare three FCN-based networks (two existing models and a new proposed in this paper for comparison) on multiple datasets. It has been observed in practice that the connections from the pre-pooled feature maps to the post-upsampled can achieve better results. We investigate the cause and provide experiments to shows that the skip-layer connections in FCN can promote flat optimization landscape, which is well known to generalize better. Additionally, we explore the relationship between the model’s generalization ability and loss surface under different batch sizes. Results show that large-batch training makes the model converge to sharp minimizers with chaotic vicinities while small-batch method leads the model to flat minimizers with smooth and nearly convex regions. Our work may contribute to insights and analysis for designing and training FCNs.
\end{abstract}
\section{Introduction}
It is generally difficult to solve some image processing problems without complicated procedures. For instance, in natural image edge detection, using edge detectors such as Canny operator is unable to obtain satisfactory results (Ganin and Lempitsky 2014). The textures may be extremely complex while in some cases we expect models to output specific edges instead of all in the whole image. 

In fact, many of these problems can be treated as mapping an input image to a corresponding output image. Different fully convolutional networks (FCN) have achieved great success in image-to-image tasks and enable them to be tackled end-to-end. The concept of FCN is first introduced for pixel-to-pixel semantic segmentation by removing fully connected layers and inserting upsampling layers at the end of networks (Long, Shelhamer and Darrell 2015). 

With this approach, many classification networks can be converted for image-to-image mapping (Xie and Tu 2015; Liu et al. 2017). They use upsampling layers to recover from feature maps with different resolutions and a fusion layer to output the final image. These models show good performance in edge detection of natural images. Apart from modifying classification networks, (Ronneberger, Fischer and Brox 2015) design an encoder-decoder structure consisting of a contracting path and multiple expanding paths to capture context at different scales. It has been applied for biomedical image segmentation successfully.

Trying and training different architectures empirically can obtain satisfactory results but clear understandings of the improvement in performance are insufficient. Theoretically exploring loss surfaces of classification neural networks have been proposed recently (Nguyen and Hein 2018; Liang et al. 2018). (Li et al. 2017) shows that the residual mapping structure can lead to a convex objective function using a visualization method. However, there is no similar analysis on FCNs, which have totally different structures and objective functions from classification models. Therefore, we make an exploration on FCN-based networks by choosing a solution and projecting its vicinity along two random vectors. To the best of our knowledge, our paper is the first to provide a visualized insight into the minimizers’ geometry of FCNs and their generalization abilities. 

The rest of this paper is organized as follows. Firstly, representative FCN-based models and visualization approaches are discussed. Then we introduce the objective function, visualization method as well as different architectures used for comparison in this study. Thirdly, we evaluate these FCNs on multiple datasets and visualize the optimization landscapes to understand their generalization abilities. Detailed analysis is also provided. Finally, we conclude the paper. 

\section{Related Work}
\subsection{Full Convolutional Networks}
Fully convolutional network (FCN) is first proposed in (Long, Shelhamer and Darrell 2015), which makes it possible for image classification networks to output segmentation results. In general, fully connected layers are applied at the end of CNN to predict class scores. Features maps will be flattened and fed into classifier layers. Space information is lost due to this squeezing operation. They replace the fully connected layers with upsampling or de-convolution layers. Therefore, image classification models can operate inputs of any size and generate corresponding output images.

Many basic classification networks, such as VGG-Net (Simonyan and Zisserman 2015), can be used as a backbone for learning image mapping by leveraging techniques in (Long, Shelhamer and Darrell 2015). Holistically-nested Edge Detector (HED) extracts features with different resolutions from five layers of VGG-Net and upsamples them to original scales (Xie and Tu 2015). At the end of the model, it uses a fusion layer to concatenate the features maps in a given dimension and output a corresponding result. Besides both SegNet and DeconvNet have a structure consisting of an encoder network and a decoder network, where the encoder topology is identical to VGG-Net (Noh, Hong and Han. 2015; Vijay, Alex and Roberto 2017). These two models learn to do upsampling using transposed convolution and then convolve with kernels to produce feature maps for segmentation.

Instead of modifying classification models, another type of networks consist of a contracting path to capture context and multiple symmetric expanding paths that enable precise localization. U-Net is one of the most representative models, which is initially designed for biomedical image segmentation and won the ISBI cell tracking challenge in 2015 by a large margin (Ronneberger, Fischer and Brox 2015). Further variations of U-Net are proposed to solve different tasks in the literature (Milletari, Navab and Ahmadi 2016; Litjens et al. 2017). (Santhanam, Morariu and Davis 2017) creates a similar framework and achieves good performance on relighting, denoising and colorization.
 
Unlike the former methods, (Zhang et al. 2017) designs an FCN that removes all pooling layers and keeps the image size inside for image denoising. Batch normalization and residual learning are also integrated to boost the training process as well as the performance. They achieve decent denoising results, though this architecture may have relatively smaller receptive field and higher computing resources consumption.
\subsection{Visualization Methods}
Although a large number of FCNs have been designed for various tasks, visualized methods are only applied on classification networks. It is well known that the objective function of deep neural networks lies in a super high dimensional space. As a result, we can only visualize it in 2D or 3D space. 

A widely-used 2D plotting method is linear interpolation, which is first applied to study local minimizers under different batch size settings (Dinh et al. 2017; Keskar et al. 2017). An alternate approach is projection. (Goodfellow, Vinyals and Saxe 2015) uses this approach to explore the trajectories of different optimization algorithms. (Li et al. 2017) studies the relationship between loss landscapes and models’ generalization abilities by projection. Note that these studies only focus on classification networks, whose structures and objective functions are entirely different from FCNs. 

\section{Method}
\subsection{Objective Function}
Image-to-image mapping problems can be converted as pixel regression or classification using FCNs. The main difference is whether the output is continuous or categorical. 

We train networks as a regression problem since pixel-wise class labels are slightly difficult to obtain in practice. For instance, the raindrop removal dataset only provides original images as labels. Therefore, we use mean square error (MSE) as the objective function, which can be expressed as follows:
\begin{equation}
L=\frac{1}{N}\sum_{i=1}^{N}\sum_{t=1}^{M}(y^{i}_{t}-\hat{y}^{i}_{t})
\label{eq1}\end{equation}
where $N$ is used to represent the number of samples and $M$ denotes the dimension of outputs. Generally, the $L1$ loss and MSE loss are almost the same in terms of regression accuracy, while sometimes the result of MSE is slightly better (Zhang et al. 2016).
\subsection{Visualization of Optimization Landscapes}
By following (Goodfellow, Vinyals and Saxe 2015; Li et al. 2017), we choose a solution $\theta^{'}$
of the network with parameters $\theta$ and sample two set of random vectors $u = \{u_1, u_2,…, u_{|\theta|}\}$and $v = \{v_1, v_2,…, v_{|\theta|}\}$ from a Gaussian distribution $N(\mu, \Sigma)$, where $|\theta|$ denotes the number of filters in FCNs, $\mu$ is set at zero and $\Sigma$ is assumed to be the identity matrix $I$. After that, the optimization landscape can be plotted by projecting the vicinity of  $\theta^{'}$ onto a 3D space along these two random directions. It can be formulated as the following equation:
\begin{equation}
f(\alpha,\beta)=L(\theta^{'}+\alpha \frac{u}{\|u\|} + \beta \frac{v}{\|v\|})
\label{eq2}\end{equation}
where $\alpha$ and $\beta$ determine the optimization step size. We do filter-wise normalization by dividing the $L2$ norm of each sampled vector $u_i$ and $v_i$. For better understanding our method, the pseudocode of visualization algorithm is provided in Algorithm 1.

\begin{algorithm}[htb] 
\caption{Visualize Loss Function $L$ } 
\label{alg:Framwork} 
\begin{algorithmic}[1] %这个1 表示每一行都显示数字
\REQUIRE ~~
A solution $\theta^{'}$of the network, plotting resolution $n$ and a visualization interval $[-r, r]$
\ENSURE ~~
A $n\times n$ loss matrix $F$
\WHILE {$i = 1, 2, …, |\theta|$}
\STATE  sample $u_i$ from $N(0, I_i)$, where $I_i$ has the same dimension as $i$-th filter $\theta_i$ of the network
\STATE sample $v_i$  from $N(0,  I_i)$
\STATE $u_i=\frac{u_i}{\|u_i\|}$
\STATE $v_i=\frac{v_i}{\|v_i\|}$
\ENDWHILE
\WHILE {$ t = 0, 1, …, n $}
\STATE $\alpha_t = -r + 2r\frac{t}{n}$
\WHILE{$ k = 0, 1, …, n $}
\STATE $\beta_k = -r + 2r\frac{k}{n}$
\STATE $\theta_{new} = \emptyset$
\FOR{$i = 1, 2, …, |\theta|$}
\STATE $\theta_{i} = \theta_{i}^{'}+\alpha_t u_i + \beta_k v_i$
\STATE $\theta_{new} = \theta_{new} \cup \theta_{i} $
\ENDFOR
\STATE $F_{t,k} = L(\theta_{new})$
\ENDWHILE
\ENDWHILE
\RETURN $F$
\end{algorithmic}
\end{algorithm}

There may be two ambiguities about this approach. One is probably about the generalization of local minimizers. Although finding a global minimum turns out to be an NP-hard problem for the non-convex objective function of deep neural networks, (Kawaguchi 2016; Lu and Kawaguchi 2017) show that local minima is not an issue for networks to generalize well. Another consideration may be the repeatability of 3D projection plot using random vectors for the high-dimension loss functions. (Li et al. 2017) has shown that different random directions can produce similar shape by plotting multiple times. This phenomenon also exists in our experiments.

\subsection{Fully Convolutional Networks}
We compare three fully convolutional networks including two existing representative models (FCN-16s, U-Net) and a new proposed. Figure 1 shows the brief illustrations of these three structures. 

\begin{figure}[!t]
\centerline{\includegraphics[width=\columnwidth]{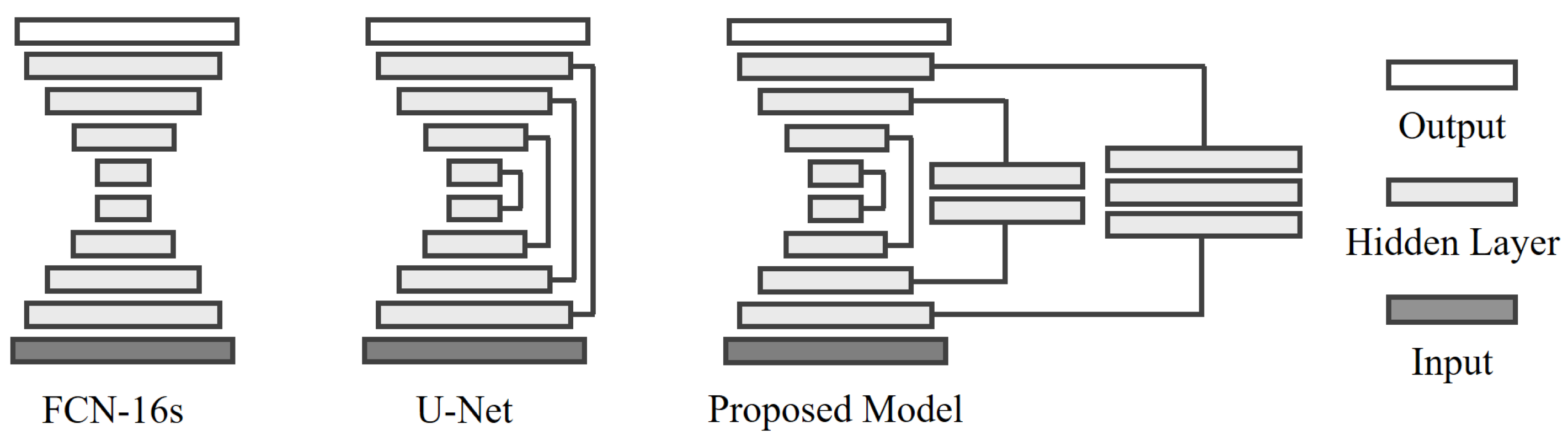}}
\caption{Abstract illustrations of three FCNs compared in our work. The channel settings of U-Net keep unchanged. FCN-16s has the same settings. We reduce the channels at low-resolution layers correspondingly in the proposed model to keep the number of parameters in the same order of magnitude.}
\label{fig1}
\end{figure}

\subsubsection{FCN-16s} Here FCN-16s refers to the network proposed in (Long, Shelhamer and Darrell 2015). For better comparison, we slight adjust it to have an almost same encoder-decoder structure as U-Net, except that the fusion is only from the last max pooling layer.

\subsubsection{U-Net} U-Net is the model designed in (Ronneberger, Fischer and Brox 2015). It is set up as follows: Convolutional and max pooling layers are stacked to generate feature maps with different resolutions. Before each max pooling process, the network branches off and connects to the input of every upsampling layer. This encoder-decoder structure has been widely used as an entire network or a key component in image-mapping tasks.

\subsubsection{Proposed Network} The network in the right plot of Figure 1 is proposed in this paper for comparison. The design of this model is motivated by the need not only to capture global context at every scale but consolidate the local information at original resolution as well. Our model inherits the main backbone from U-Net. The differences between our model and U-Net is that we insert a various number of residual modules (He et al. 2016) between the pre-pooled and up-sampled feature maps to emphasize the importance of information from original resolution. We do not use a convolution layer with kernel size greater than 7, since using smaller kernels can still capture a large spatial context (Szegedy et al. 2015). We note that HourglassNet has similar skip connections with convolutional layers inserted (Newell, Yang and Deng 2016), but is different from this model.

\section{Experiments}
\subsection{Experimental Datasets and Settings}
To compare FCN-based networks, we conduct different image-to-image experiments including electron microscope (EM) imaging segmentation, vessel extraction and raindrop removal. Descriptions of datasets, models, training details and evaluate methods are provided respectively. Unless otherwise mentioned, we split each dataset as training and testing set according to 7:3. All images in training sets are flipped and rotated at every 90 degrees. Eight times data are obtained after augmentation. All models are trained using the SGD optimizer with batch-size 16 and momentum 0.8 for 60 epochs. The learning rate is initialized at 0.025. Note that our goal is not to achieve state-of-the-art performance on these tasks but rather to study generalization abilities of FCNs through visualizing the optimization landscapes.

\subsubsection{EM Segmentation} The dataset is provided by the EM segmentation challenge (Arganda-Carreras et al. 2015). In our experiments, each image is cropped as several 256$\times$256 patches with a 128-pixel overlap. Foreground-restricted rand score and information theoretic score after border thinning are used to evaluate the quality of segmentation results (Unnikrishnan, Pantofaru and Hebert 2007).
\subsubsection{Vessel Extraction}   Experiments are implemented on the Digital Retinal Images for Vessel Extraction database (DRIVE) proposed for studies on the extraction of blood vessels (Staal et al. 2004). Models are trained after reshaping all data at 256$\times$256 size. Rand score and information theoretic score is also used as evaluation metrics.
\subsubsection{Raindrop Removal}   We conduct raindrop removal experiments on the database created by (Qian et al. 2018). There are 1100 pairs of images totally with various background scenes and raindrops. We resize all images to 480$\times$480 for training. Peak signal-to-noise ratio (PSNR) and structure similarity index (SSIM) are applied for quantitative comparison.

\subsection{Results on Image-to-image Mapping Tasks}
The following provides quantitative comparisons of three FCNs respectively. Example results on various image-to-image mapping tasks are shown in Figure 2.

\begin{figure*}[ht] 
\centering
\subfigure{\includegraphics[width=0.19\linewidth]{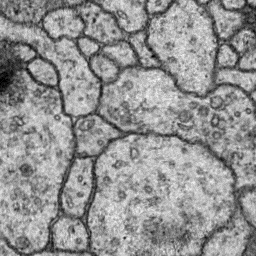}}
\subfigure{\includegraphics[width=0.19\linewidth]{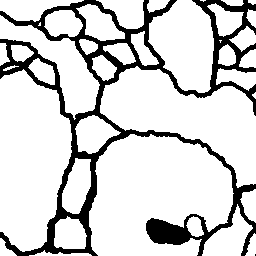}}
\subfigure{\includegraphics[width=0.19\linewidth]{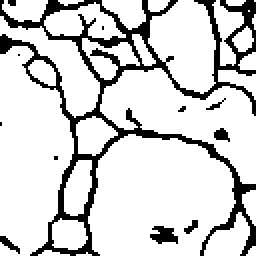}}
\subfigure{\includegraphics[width=0.19\linewidth]{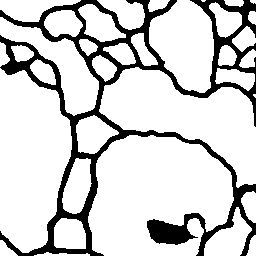}}
\subfigure{\includegraphics[width=0.19\linewidth]{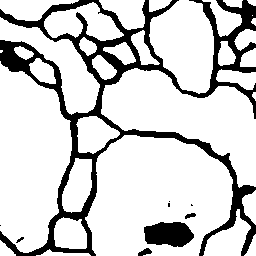}}
\subfigure{\includegraphics[width=0.19\linewidth]{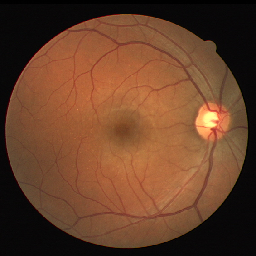}}
\subfigure{\includegraphics[width=0.19\linewidth]{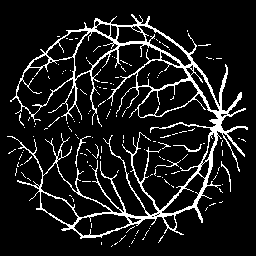}}
\subfigure{\includegraphics[width=0.19\linewidth]{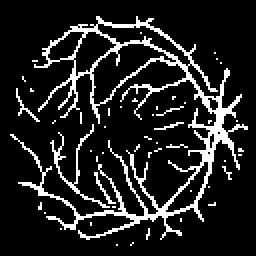}}
\subfigure{\includegraphics[width=0.19\linewidth]{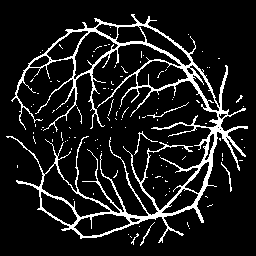}}
\subfigure{\includegraphics[width=0.19\linewidth]{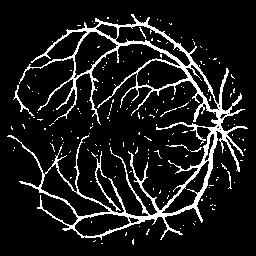}}
\subfigure{\includegraphics[width=0.19\linewidth]{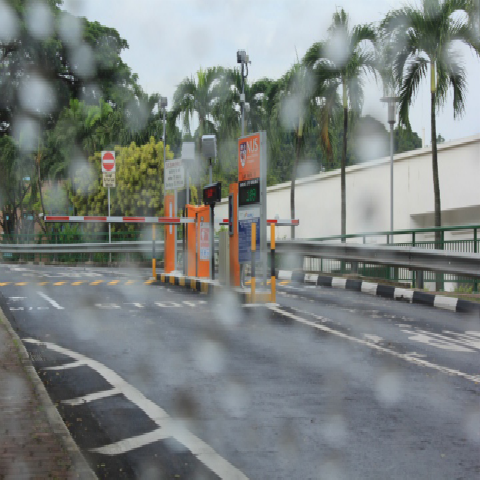}}
\subfigure{\includegraphics[width=0.19\linewidth]{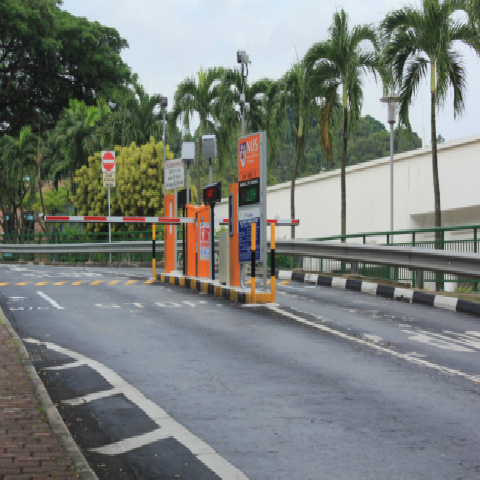}}
\subfigure{\includegraphics[width=0.19\linewidth]{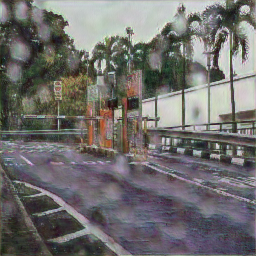}}
\subfigure{\includegraphics[width=0.19\linewidth]{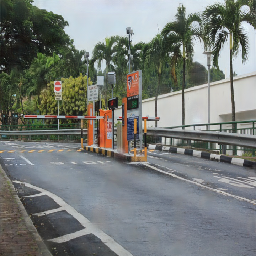}}
\subfigure{\includegraphics[width=0.19\linewidth]{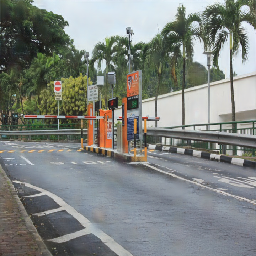}}
\caption{Example results of three FCNs (from left to right: input, ground truth, the output of FCN-16s, U-Net and our model) on various tasks (from top to bottom: EM segmentation, vessel extraction and raindrop removal)}
\label{fig2} 
\end{figure*}

\subsubsection{EM segmentation}
We plot the training and testing loss curves during the 60-epoch period in Figure 3 aiming to compare the training efficiency and generalization ability of the three networks. It also shows that the difference in generalization is not caused by underfitting or overfitting. As can be seen, U-Net and FCN-16s converge at the nearly same speed and faster than our model. In contrast, the testing loss reveals that U-Net generalizes better than FCN-16s on the EM segmentation task. Our model shows  better generalization than FCN-16s. Interestingly there is a fluctuation occurs in the testing loss, for which we provide explanations in visualization experiments. Table 1 lists the best quantitative results during these 60 epochs, which is consistent with the testing MSE loss.

\begin{figure} 
\centering
\subfigure{\includegraphics[width=0.49\linewidth]{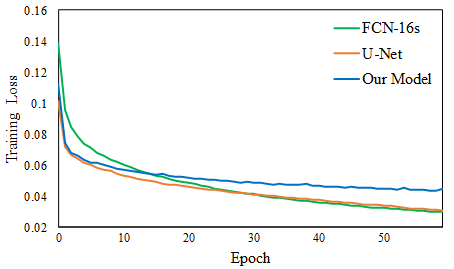}}
\subfigure{\includegraphics[width=0.49\linewidth]{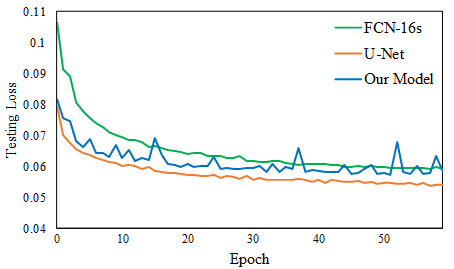}}
\caption{Comparison of the training efficiency and generalization ability of three FCNs on EM segmentation.}
\label{fig3} 
\end{figure}

\begin{table}
\caption{Quantitative results on EM segmentation}
\label{tab1}
\begin{tabular}{ccc}
\hline
Model & Rand Score & Information Theoretic Score\\
\hline
FCN-16s & 0.5697 & 0.5786\\
U-Net & 0.9714 & 0.9743\\
Our Model & 0.9296 & 0.9407\\
\hline
\end{tabular}
\end{table}

\subsubsection{Vessel Extraction}
Table 2 lists the best evaluation results. It shows that our model performs on par with U-Net and better than FCN-16s.

\begin{table}
\caption{Quantitative results on DRIVE testing set}
\label{tab2}
\begin{tabular}{ccc}
\hline
Model & Rand Score & Information Theoretic Score\\
\hline
FCN-16s & 0.6594 & 0.6667\\
U-Net & 0.7832 & 0.7974\\
Our Model & 0.7696 & 0.7894\\
\hline
\end{tabular}
\end{table}

\subsubsection{Raindrop Removal}
   The average PSNR and SSMI results on raindrop removal are presented in Table 3. It can be seen that U-Net yields the highest PSNR and SSMI on this task and our model achieves better performance than FCN-16s. 

\begin{table}
\caption{ Average PSNR and SSMI of different models for raindrop removal}
\label{tab3}
\centering
\begin{tabular}{ccc}
\hline
Model & PSNR & SSMI\\
\hline
FCN-16s & 17.043 & 0.820\\
U-Net & 27.049 & 0.982\\
Our Model & 26.328 & 0.978\\
\hline
\end{tabular}
\end{table}

\subsection{Visualization of Optimization Landscapes}
We implement the visualization of optimization landscapes on EM segmentation dataset to address the following three issues: 
\begin{itemize}
\item As illustrated in experimental results, our model and U-Net outperform on all image-to-image tasks. This phenomenon motivates us to think what the optimization landscapes of the three FCNs look like.
\item Although FCN-16s has a similar encoder-decoder structure and even the same number of parameters as U-Net, a significant difference of the two structures is that FCN-16s only preserves the feature representations after 16$\times$ max pooling. One might wonder how the skip-layer connections affect the loss surface of FCNs.
\item Previous literature has shown that for classification models small-batch training can obtain flat minimizers which generalize better than large-batch method (Keskar et al. 2017; Li et al. 2017). However, FCNs for image-to-image tasks have different structures and objective functions. Does this phenomenon still exist in FCNs? If it exists, what the difference of minimizers that are converged with small/large-batch training methods? 
\end{itemize}

\subsubsection{Optimization Landscapes of the Three FCNs} 
 Given the computation cost in visualization, we can only plot the loss surface in low-resolution, i.e., $n = 40$ in Algorithm 1. For a better view of the optimization landscapes, we choose an interval $[-0.5, 0.5]$, i.e., $r = 0.5$. The resolution is not high enough to capture the complex non-convexity of networks in large regions. For the convenience of explanation, we first illustrate flat/sharp minimizers loosely as shown in Figure 4.

\begin{figure}[!t]
\centerline{\includegraphics[width=0.34\columnwidth]{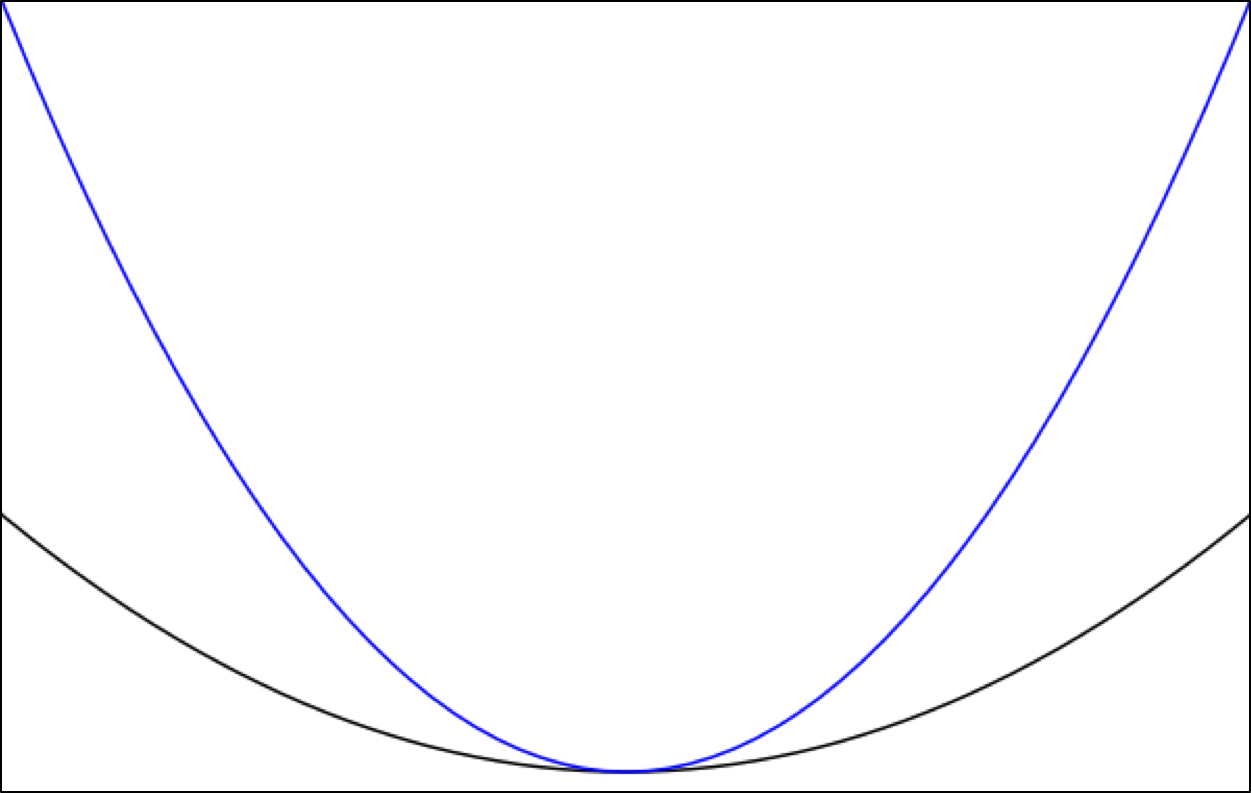}}
\caption{A conceptual illustration of the flat and sharp minimizer. The blue shows the sharp minima and the black sketches the flat.}
\label{fig4}
\end{figure}

Using the method described in the previous section, we first choose a solution of three models where the training loss value is smallest during the 60 epochs. After that, we plot the 3D loss landscapes and projected 2D contours. As shown in Figure 5, a noticeable difference is that the optimization landscape around the minimizer of FCN-16s is highly sharper while those of U-Net and our model are much flatter. Note that U-Net and FCN-16s are plotted by choosing a center point with a same training loss value.

\begin{figure*} 
\centering
\subfigure{\includegraphics[height=0.22\linewidth]{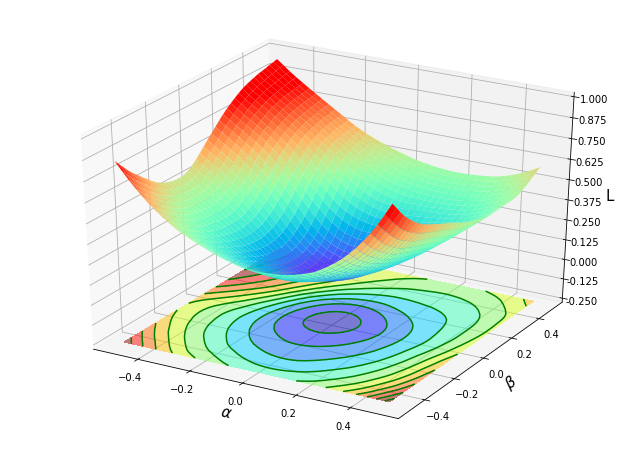}}
\subfigure{\includegraphics[height=0.22\linewidth]{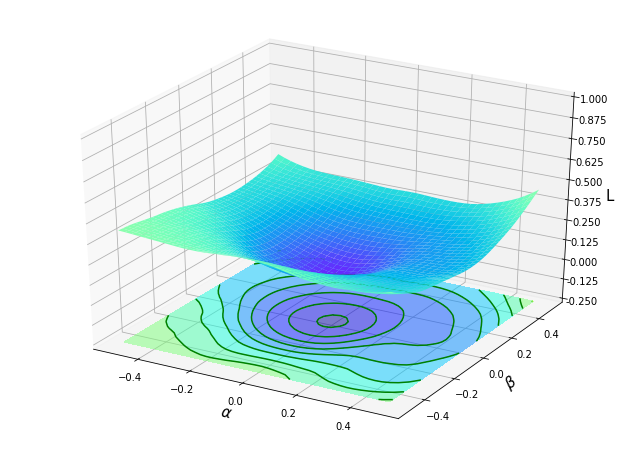}}
\subfigure{\includegraphics[height=0.22\linewidth]{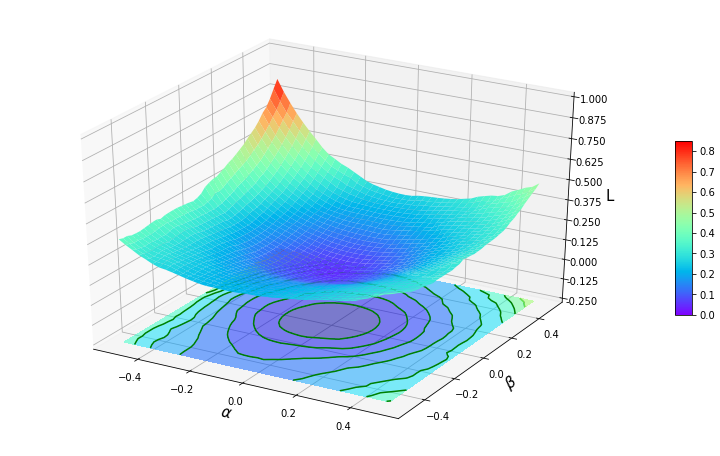}}
\caption{The 3D optimization landscape and projected 2D contour of FCNs (from left to right: FCN-16s, U-Net and our model) on EM segmentation datasets.}
\label{fig5} 
\end{figure*}

In addition to visual comparisons, the flatness/sharpness of minimizers can be described by the Hessian matrix $\nabla^2L$. However, it has extremely heavy computation burden when applying in neural networks. Hence, we adopt a metric from (Keskar et al. 2017) to characterize the sharpness.

Given $\theta \in \mathbb{R}^m$ and $\epsilon > 0$, the $C_{\epsilon}$-sharpness of $L$ at $\theta$ is defined as:
\begin{equation}
\phi_{\theta,L} = \frac{max_{\sigma \in C_{\epsilon}}L(\theta+\sigma) - L(\theta)}{1+L(\theta)}
\label{eq3}\end{equation}
where $\epsilon$ determines the size of box around the solution $\theta$ of the loss function $L$ and $C_{\epsilon}$ denotes a constraint set, which is defined as:
\begin{equation}
\begin{aligned}
C_{\epsilon} = \{z \in \mathbb{R}^m: -\epsilon(|\theta_i|+1) \le z_i \le \epsilon(|\theta_i|+1), \\
\forall i \in\{1,2,...,m\}\}
\end{aligned}
\label{eq4}\end{equation}
The operation of adding one in (3) and (4) is to guard against the zero case, which is discarded in our experiments.

We conduct the experiments with same training loss (0.03 for FCN-16s and U-Net) for 5 times and list the average sharpness value within two different constraint sets in Table 4. The smaller value sketches the flatter minimizer. As can be seen, U-Net and our model converge to flat minimizers while FCN-16s tends to achieve sharper one. The quantitative results are in alignment with our observations.

\begin{table}
\caption{Average $C_\epsilon$-sharpness of different models}
\label{tab4}
\centering
\begin{tabular}{ccc}
\hline
Model & $\epsilon = 0.1$ & $\epsilon = 0.2$ \\
\hline
FCN-16s & 2.1822 & 7.3241\\
U-Net & 1.4608 & 4.7711\\
Our Model & 1.5519 & 5.0324\\
\hline
\end{tabular}
\end{table}

\begin{figure} 
\centering
\subfigure{\includegraphics[height=0.32\linewidth]{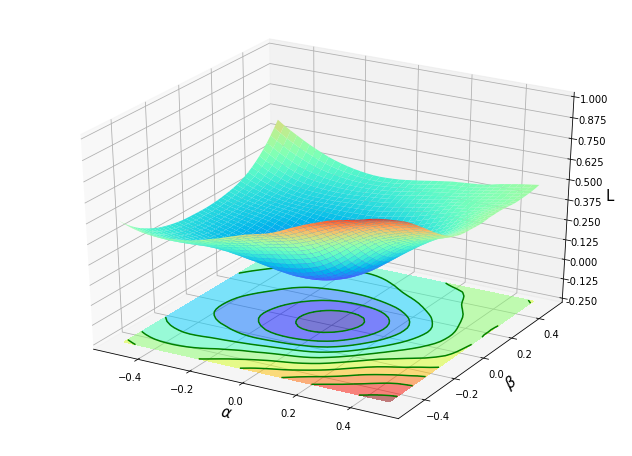}}
\subfigure{\includegraphics[height=0.32\linewidth]{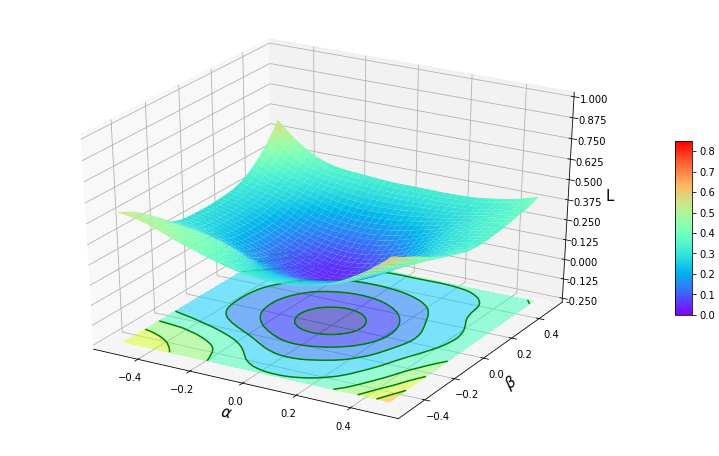}}
\caption{The 3D optimization landscape and projected 2D contour of FCN-8s (left) and FCN-4s (right) on EM segmentation dataset. Comparing together with FCN-16s (testing loss 0.05889), they are plotted at the same training loss value 0.03 but have different testing loss value (0.05524 and 0.05361).}
\label{fig6} 
\end{figure}

In fact, it has been widely discussed that convergence to sharp minimizers gives rise to the poor generalization for deep learning. The large sensitivity of the objective function at a sharp minimizer has a negative impact on the ability of the trained model to generalize on new testing data (Keskar et al. 2017). They also provide detailed literature review in statistics, Bayesian learning and Gibbs energy to support this view. Therefore, it may provide an explanation for why U-Net generalizes better than FCN-16s on EM segmentation.

\subsubsection{Skip-layer Connections Promote Flat Loss Landscape}
To explore the effect of skip-layer connections on the optimization landscape of FCNs, we conduct comparative experiments on three models, i.e., FCN-16s, FCN-8s, FCN-4s.  The notation -16s, -8s, -4s represents the skip-layer connections added after 16$\times$, 8$\times$, 4$\times$ max pooling operations respectively. We use the same setting ($n = 40$ and $r = 0.5$) and report the sharpness values in Table 5. The visualization results are shown in Figure 6. We use all networks with nearly same training loss 0.03 in experiments, but there is a significant difference in the generalization performance. Comparing together with FCN-16s in Figure 5, we see that the skip-layer connections from the pre-pooled features to the post-upsampled promote FCNs to obtain flat optimization landscapes.

\begin{table}
\caption{Average $C_\epsilon$-sharpness of FCNs with different skip-layer connections}
\label{tab5}
\centering
\begin{tabular}{ccc}
\hline
Model & $\epsilon = 0.1$ & $\epsilon = 0.2$ \\
\hline
FCN-16s & 2.1822 & 7.3241\\
FCN-8s & 1.7724 & 6.1531\\
FCN-4s & 1.4904 & 4.8242\\
\hline
\end{tabular}
\end{table}

It is well known in computer vision that context and locality are a pair of trade-off. On the one hand, max pooling modules expand local receptive fields and reduce the computation cost. On the other hand, we hope the regression prediction to be pixel perfect, which is slightly difficult for upsampling layers to restore from coarser features. 
Because there is sort of heterogeneity and deterioration that happens inside these feature maps after max pooling operations. Some sharp boundaries and tiny details at original scale will loss to some extent. Therefore, it is necessary to add skip-layers to alleviate this problem. The loss landscape transition from sharp to flat possibly explains why expanding connection path at finer features is important in FCNs from the perspective of optimization.

\subsubsection{Small-batch Training Leads FCNs to Flatter Minimizers with Smooth Vicinities That Generalizes Better}
In order to answer the last question, we trained the three FCNs with small batch 2 (B2) and large batch 16 (B16). Note that FCNs generally have more computational cost, which makes it difficult to train using batch size as large (say 128-512) as classification models with limited computation resources. The training and testing loss curves are shown in Figure 7. It is obvious that models trained with large batch sizes generalize worse.
\begin{figure} 
\centering
\subfigure{\includegraphics[width=0.32\linewidth]{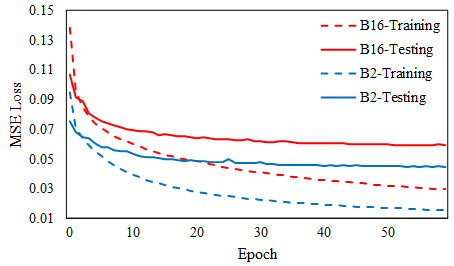}}
\subfigure{\includegraphics[width=0.32\linewidth]{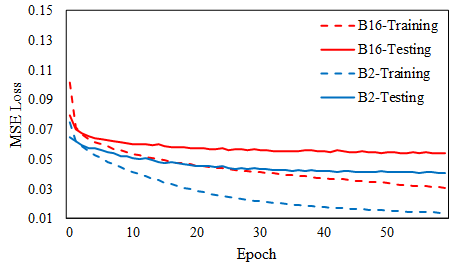}}
\subfigure{\includegraphics[width=0.32\linewidth]{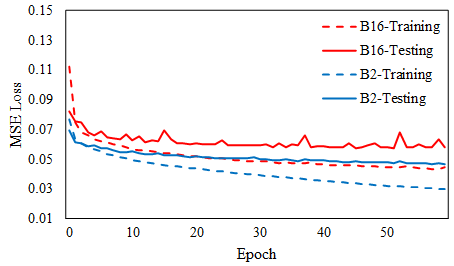}}
\caption{Training and testing loss for large-batch (B16) and small-batch (B2) method in 60 epochs (from left to right: FCN-16s, U-Net and our model). }
\label{fig7} 
\end{figure}

\begin{figure} 
\centering
\subfigure{\includegraphics[height=0.32\linewidth]{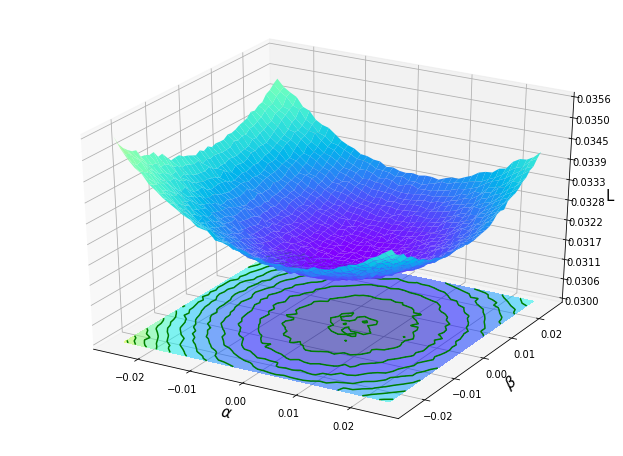}}
\subfigure{\includegraphics[height=0.32\linewidth]{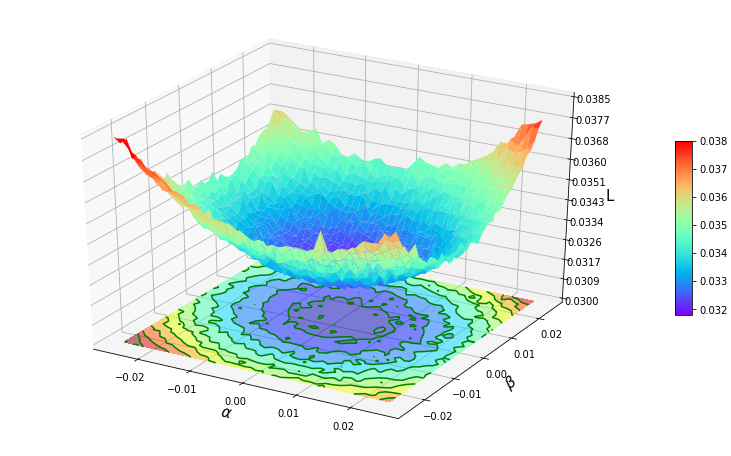}}\\
\subfigure{\includegraphics[height=0.32\linewidth]{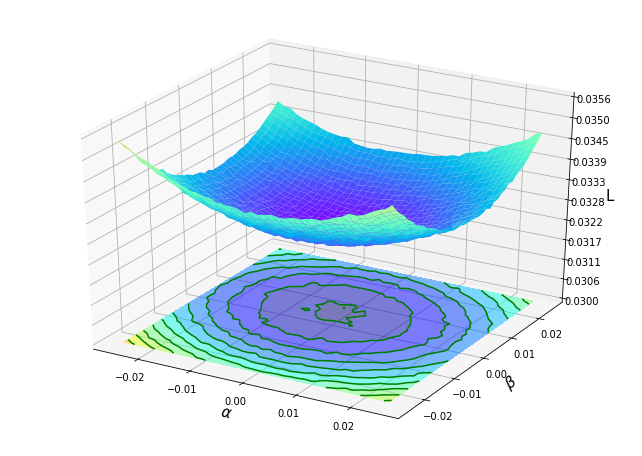}}
\subfigure{\includegraphics[height=0.32\linewidth]{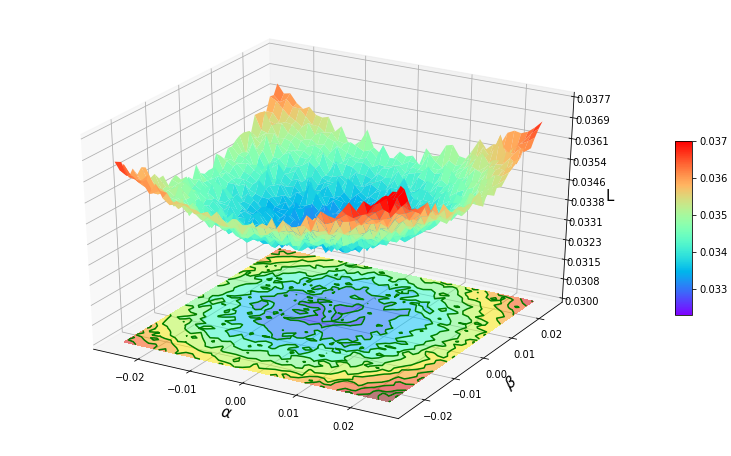}}\\
\subfigure{\includegraphics[height=0.32\linewidth]{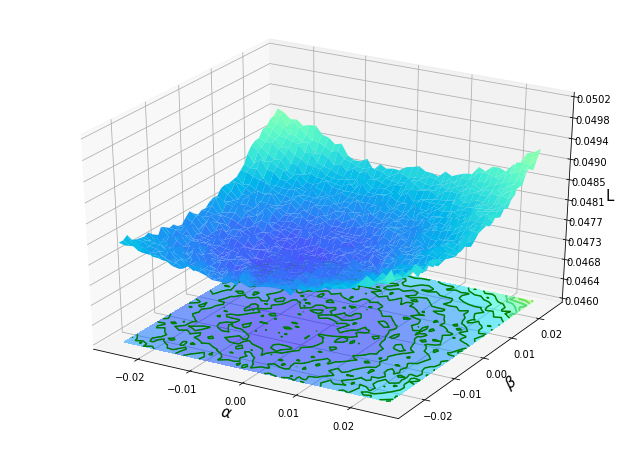}}
\subfigure{\includegraphics[height=0.32\linewidth]{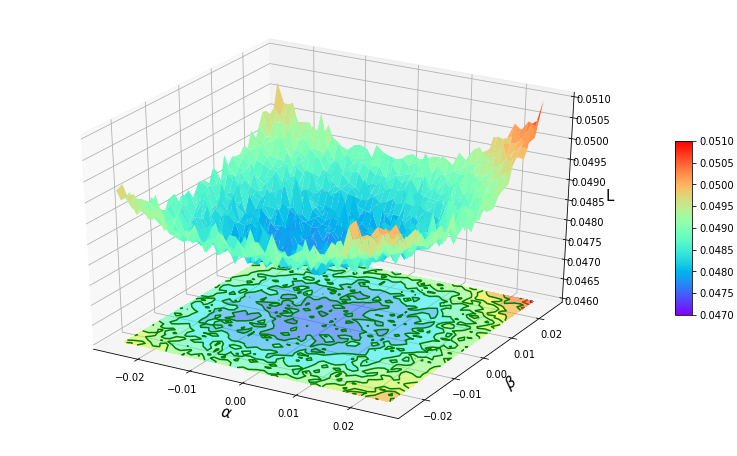}}
\caption{ Side-by-side comparison for small-batch and large-batch method (from left to right: B2 and B16; from top to bottom: FCN-16s, U-Net and our model. The same model is plotted at the same training loss).}
\label{fig8} 
\end{figure}

To investigate the reasons, we visualize their optimization landscapes using the same resolution ($n = 40$) but smaller region ($r = 0.025$). This time we compare the minimizer’s flatness of the same network. As shown in Figure 8, small-batch method leads the model to flat minimizers with smooth and nearly convex regions while large-batch training makes the model converge to sharp minimizers with chaotic vicinity. Table 6 lists the sharpness values.

\begin{table}
\centering
\caption{Average $C_\epsilon$-sharpness of FCNs with small/large-batch training}
\label{tab6}
\begin{tabular}{ccccc}
\hline
\multirow{2}{*}{Model}&
\multicolumn{2}{c}{$\epsilon = 0.01$}&
\multicolumn{2}{c}{$\epsilon = 0.02$}\\
\cline{2-5}
  & B2&B16 & B2&B16\\
\hline
FCN-16s & 0.0252 & 0.0418 & 0.1193 & 0.1964\\
U-Net & 0.0175 & 0.0288 & 0.0971 & 0.1344\\
Our Model & 0.0187 & 0.0315 & 0.1072 & 0.1699\\
\hline
\end{tabular}
\end{table}

\section{Conclusion}
In conclusion, visualized insights into the optimization landscapes of FCNs are provided. We observe that consolidating the original scale features by adding skip-layer connections in FCNs can promote flat loss landscape, which is well related to good generalization. In addition, experiments are conducted to show models trained with small batch sizes generalize better. We investigate the cause and present evidence that small-batch method leads the model to flat minimizers with smooth and nearly convex regions while large-batch training makes the model converge to sharp minimizers with chaotic vicinities. Admittedly theoretical proof is difficult to provide at present, but our work may contribute to understanding and analysis for designing and training FCNs.

\section{References} 
\smallskip \noindent Arganda-Carreras, I.; Turaga, S. C.; Berger, D. R.; and et al. 2015. Crowdsourcing the Creation of Image Segmentation Algorithms for Connectomics. \textit{Frontiers in neuroanatomy} 9(142):1-13.

\smallskip \noindent Dinh, L.; Pascanu, R.; Bengio, S.; and Bengio, Y. 2017. Sharp Minima Can Generalize for Deep Nets. In \textit{Proceedings of the 34th International Conference on Machine Learning,} 70:1019-1028.

\smallskip \noindent Ganin, Y., and Lempitsky, V. 2014. N4-Fields: Neural Network Nearest Neighbor Fields for Image Transforms. In \textit{Proceedings of Asian Conference on Computer Vision,} 536-551. Cham: Springer.

\smallskip \noindent Goodfellow, I. J.; Vinyals, O.; and Saxe, A. M. 2015. Qualitatively Characterizing Neural Network Optimization Problems. In \textit{International Conference on Learning Representations.}

\smallskip \noindent He, K.; Zhang, X.; Ren, S.; and Sun, J. 2016. Deep Residual Learning for Image Recognition. In \textit{Proceedings of the IEEE Conference on Computer Vision and Pattern Recognition,} 770-778.

\smallskip \noindent Kawaguchi, K. 2016. Deep Learning without Poor Local Minima. In \textit{Advances in Neural Information Processing Systems,} 586-594.

\smallskip \noindent Keskar, N. S.; Mudigere, D.; Nocedal, J; and et al. 2017. On Large-batch Training for Deep Learning: Generalization Gap and Sharp Minima. In \textit{International Conference on Learning Representations}.

\smallskip \noindent Li, H.; Xu, Z.; Taylor, G.; and Goldstein, T. 2017. Visualizing the Loss Landscape of Neural Nets. \textit{arXiv:1712.09913.}

\smallskip \noindent Li, Y.; and Yuan, Y. 2017. Convergence Analysis of Two-layer Neural Networks with ReLU Activation. In \textit{Advances in Neural Information Processing Systems,} 597-607.

\smallskip \noindent Liang, S.; Sun, R.; Li, Y.; and Srikant, R. 2018. Understanding the Loss Surface of Neural Networks for Binary Classification. \textit{arXiv:1803.00909.}

\smallskip \noindent Litjens, G.; Kooi, T.; Bejnordi, B. E.; and et al. 2017. A Survey on Deep Learning in Medical Image Analysis. \textit{Medical Image Analysis }42: 60-88.

\smallskip \noindent Liu, Y.; Cheng, M. M.; Hu, X.; Wang, K.; and Bai, X. 2017. Richer Convolutional Features for Edge Detection. In \textit{Proceedings of the IEEE Conference Computer Vision and Pattern Recognition,} 5872-5881.

\smallskip \noindent Long, J.; Shelhamer, E.; and Darrell, T. 2015. Fully Convolutional Networks for Semantic Segmentation. In \textit{Proceedings of the IEEE Conference on Computer Vision and Pattern Recognition,} 3431-3440.

\smallskip \noindent Lu, H.; and Kawaguchi, K. 2017. Depth Creates No Bad Local Minima. \textit{arXiv:1702.08580}.

\smallskip \noindent Milletari, F.; Navab, N.; xand Ahmadi, S. A. 2016. V-Net: Fully Convolutional Neural Networks for Volumetric Medical Image Segmentation. In \textit{IEEE International Conference on 3D Vision,} 565-571.

\smallskip \noindent Newell, A.; Yang, K.; and Deng, J. 2016. Stacked Hourglass Networks for Human Pose Estimation. In \textit{Proceedings of European Conference on Computer Vision,} 483-499. Cham: Springer.

\smallskip \noindent Nguyen, Q., and Hein, M. 2018. Optimization Landscape and Expressivity of Deep CNNs. In \textit{International Conference on Machine Learning,} 3727-3736.

\smallskip \noindent Noh, H.; Hong, S.; and Han, B. 2015. Learning Deconvolution Network for Semantic Segmentation. In \textit{Proceedings of the IEEE International Conference on Computer Vision,} 1520-1528.

\smallskip \noindent Qian, R.; Tan, R. T.; Yang, W.; Su, J.; and Liu, J. 2018. Attentive Generative Adversarial Network for Raindrop Removal from a Single Image. In \textit{European Conference on Computer Vision,} 694-711. Cham: Springer.

\smallskip \noindent Ronneberger, O.; Fischer, P.; and Brox, T. 2015. U-Net: Convolutional Networks for Biomedical Image Segmentation. In \textit{International Conference on Medical Image Computing and Computer-assisted Intervention,} 234-241. Cham: Springer.

\smallskip \noindent Santhanam, V.; Morariu, V. I.; and Davis, L. S. 2017. Generalized Deep Image to Image Regression. In \textit{Proceedings of the IEEE Conference on Computer Vision and Pattern Recognition,} 5609-5619.

\smallskip \noindent Simonyan, K., and Zisserman, A. 2015. Very Deep Convolutional
Networks for Large-scale Image Recognition. In \textit{International Conference on Learning Representations}.

\smallskip \noindent Staal, J.; Abràmoff, M. D.; Niemeijer, M.; and et al. 2004. Ridge-based Vessel Segmentation in Color Images of the Retina. \textit{IEEE Transactions on Medical Imaging} 23(4), 501-509.

\smallskip \noindent  Szegedy, C.; Liu, W.; Jia, Y.; and et al. 2015. Going Deeper with Convolutions. In \textit{Proceedings of the IEEE Conference on Computer Vision and Pattern Recognition,} 1-9.

\smallskip \noindent Unnikrishnan, R.; Pantofaru, C.; and Hebert, M. 2007. Toward Objective Evaluation of Image Segmentation Algorithms. \textit{IEEE Transactions on Pattern Analysis and Machine Intelligence} 6: 929-944.

\smallskip \noindent Vijay, B.; Alex K.; and Roberto C. 2017. SegNet: A Deep Convolutional Encoder-Decoder Architecture for Image Segmentation. \textit{IEEE Transactions on Pattern Analysis and Machine Intelligence} 39(12):2481-2495

\smallskip \noindent Xie, S., and Tu, Z. 2015. Holistically-nested Edge Detection. In \textit{Proceedings of the IEEE International Conference on Computer Vision,} 1395-1403.

\smallskip \noindent  Zhang, C. L.; Zhang, H.; Wei, X. S.; and Wu, J. 2016. Deep Bimodal Regression for Apparent Personality Analysis. In \textit{European Conference on Computer Vision,} 311-324. Cham: Springer.

\smallskip \noindent Zhang, K.; Zuo, W.; Chen, Y.; Meng, D.; and Zhang, L. 2017. Beyond a Gaussian Denoiser: Residual Learning of Deep CNN for Image Denoising. \textit{IEEE Transactions on Image Processing} 26(7): 3142-3155.

\end{document}